\pdfoutput=1
\documentclass[letterpaper]{article} 
\usepackage{aaai23}  
\usepackage{times}  
\usepackage{helvet}  
\usepackage{courier}  
\usepackage[hyphens]{url}  
\usepackage{graphicx} 
\usepackage{natbib}  
\usepackage{caption} 
\usepackage{algorithm}
\usepackage{algorithmic}
\usepackage{amsmath}
\usepackage{amsmath}
\usepackage{amssymb}
\usepackage{booktabs}
\usepackage{multirow}
\usepackage{nicematrix}
\usepackage{makecell}
\usepackage{nicematrix}
\usepackage{bm}

\usepackage{float}
\usepackage{subfig}

\usepackage{newfloat}
\usepackage{listings}
\DeclareCaptionStyle{ruled}{labelfont=normalfont,labelsep=colon,strut=off} 
\floatstyle{ruled}
\newfloat{listing}{tb}{lst}{}
\floatname{listing}{Listing}
\title{Lifelong Person Re-Identification via Knowledge Refreshing and Consolidation}
\author{
    Chunlin Yu\textsuperscript{\rm 1},
    Ye Shi\textsuperscript{\rm 1,3},
    Zimo Liu\textsuperscript{\rm 2},
    Shenghua Gao\textsuperscript{\rm 1,3},
    Jingya Wang\textsuperscript{\rm 1,3}\thanks{Corresponding author}\\
}
\affiliations{
    \textsuperscript{\rm 1}ShanghaiTech University \textsuperscript{\rm 2}Peng Cheng Laboratory\textsuperscript{\rm 3} \\Shanghai Engineering Research Center of Intelligent Vision and Imaging \\

    \{yuchl, shiye, gaoshh, wangjingya\}@shanghaitech.edu.cn; liuzm@pcl.ac.cn
}

\usepackage{bibentry}

\begin{document}

\maketitle

\begin{abstract}
Lifelong person re-identification (LReID) is in significant demand for real-world development as a large amount of ReID data is captured from diverse locations over time and cannot be accessed at once inherently. However, a key challenge for LReID is how to incrementally preserve old knowledge and gradually add new capabilities to the system. Unlike most existing LReID methods, which mainly focus on dealing with catastrophic forgetting, our focus is on a more challenging problem, which is, not only
trying to reduce the forgetting on old tasks but also aiming to improve the model performance on both new and old tasks during the lifelong learning process. Inspired by the biological process of human cognition where the somatosensory neocortex and the hippocampus work together in memory consolidation, we formulated a model called Knowledge Refreshing and Consolidation (KRC) that achieves both positive forward and backward transfer. More specifically, a knowledge refreshing scheme is incorporated with the knowledge rehearsal mechanism to enable bi-directional knowledge transfer by introducing a dynamic memory model and an adaptive working model. Moreover, a knowledge consolidation scheme operating on the dual space further improves model stability over the long term. Extensive evaluations show KRC's superiority over the state-of-the-art LReID methods on challenging pedestrian benchmarks. Code is available at \textit{\texttt{\color{magenta}{https://github.com/cly234/LReID-KRKC}}}.
\end{abstract}

\section{Introduction}

\;\; Person re-identification (ReID) aims to recognize the same person reappearing in different camera views and locations. Person ReID has long been investigated in surveillance systems as a person retrieval problem, where myriad efforts have been made to tackle various challenges such as occlusions and pose variations. However, the current paradigm often sidesteps the desiderata of real-world deployments to successively learn
timely-updated new tasks and retain old knowledge in an ever-changing environment. 

In particular, when ReID systems are incrementally learned on a sequence of data collected from diverse locations and domains over time, they will often fit in the current data distribution to the detriment of previously gained knowledge. As a result, \textit{catastrophic forgetting} may occur. This suboptimal result has inspired the community to seek sustainable approaches to lifelong person ReID (LReID). Compared to conventional lifelong learning, which focuses on closed-set recognition problems, LReID is inherently a fine-grained open-set problem. The identities in the training and testing sets for LReID are non-overlapped, and one of the goals of LReID is to improve the model's generalization ability among various distributions. 
\begin{figure}[t]
        \includegraphics[width=\columnwidth]{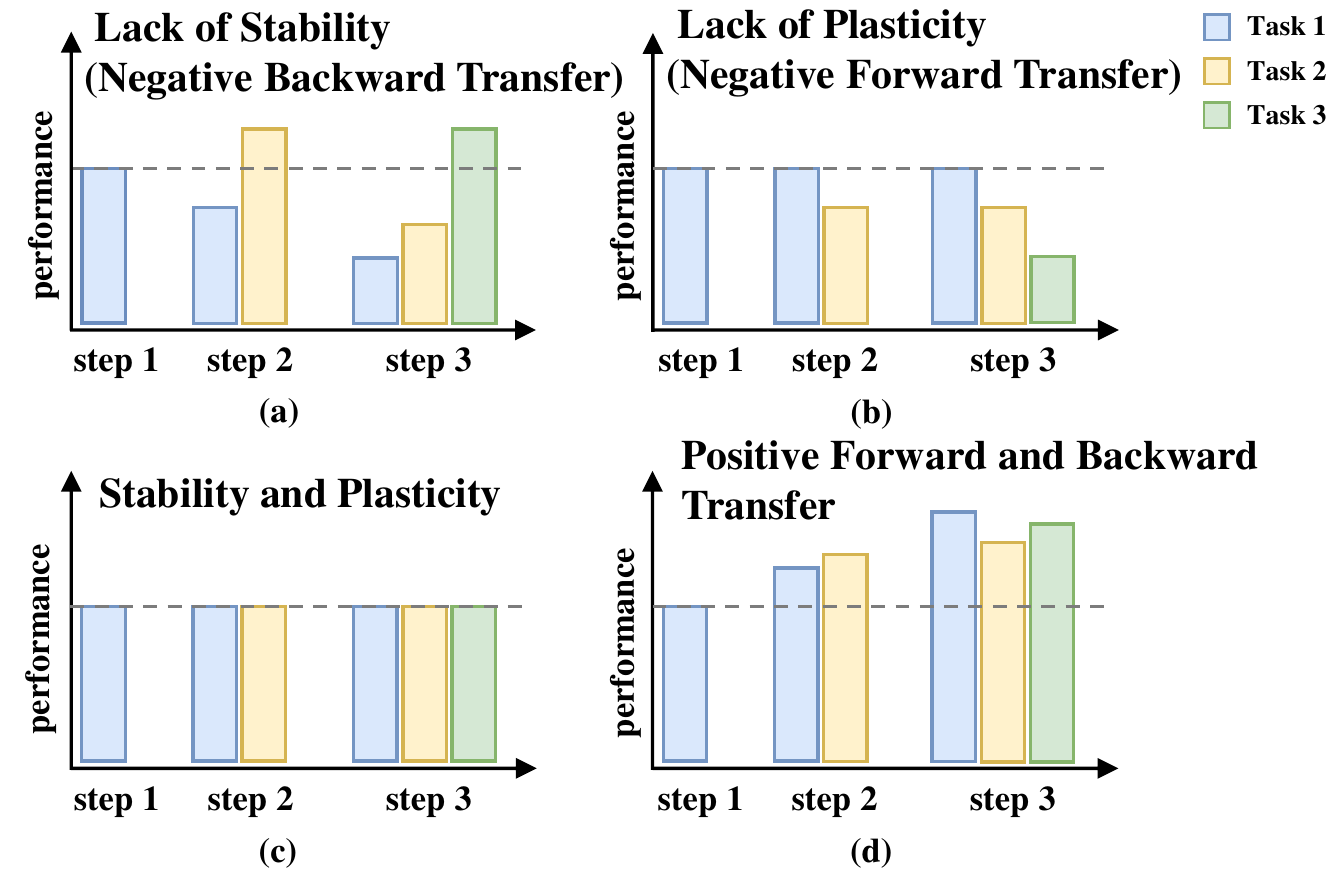}
\caption{Four scenarios of forward and backward transfer for lifelong learning. The dashed line denotes when a single task has been trained separately at random initialization.}
\label{fig1}

\end{figure}

To date, LReID has been the topic of several studies. For example, \cite{pu2021lifelong} studied knowledge accumulation, while \cite{wu2021generalising} and \cite{ge2022lifelong} both examined ways to replay selected exemplars with ad-hoc techniques. While these methods leverage
knowledge distillation or knowledge accumulation as a countermeasure
to prevent catastrophic forgetting, dealing with this issue leaves us far from solving the whole problem. Here, we target a more challenging LReID problem, which is not only trying to minimize catastrophic forgetting but also tackling the idea that model performance can be improved on both new and old tasks during the lifelong learning process. Generally, lifelong learning models can be classified into four categories according to the stability and plasticity model performance: Fig.  \ref{fig1}(a) shows that models that lack stability are prone to forgetting old tasks; Fig. \ref{fig1}(b) shows a model that has overcome the catastrophic forgetting problem but lacks plasticity to handle new tasks;  Fig.\ref{fig1}(c) shows a model that maintains good performance in terms of both stability and plasticity, however, it lacks positive knowledge transfer in both directions; Lastly, Fig. \ref{fig1}(d) represents our target model with both positive forward and backward knowledge transfer. The model in Fig. \ref{fig1}(d) not only allows previous knowledge to benefit the current task, but also improves the performance of old tasks when learning the current task. Clearly, the model in Fig. \ref{fig1}(d) is the most desirable for LReID. 

\begin{figure}[t]
        \includegraphics[width=\columnwidth]{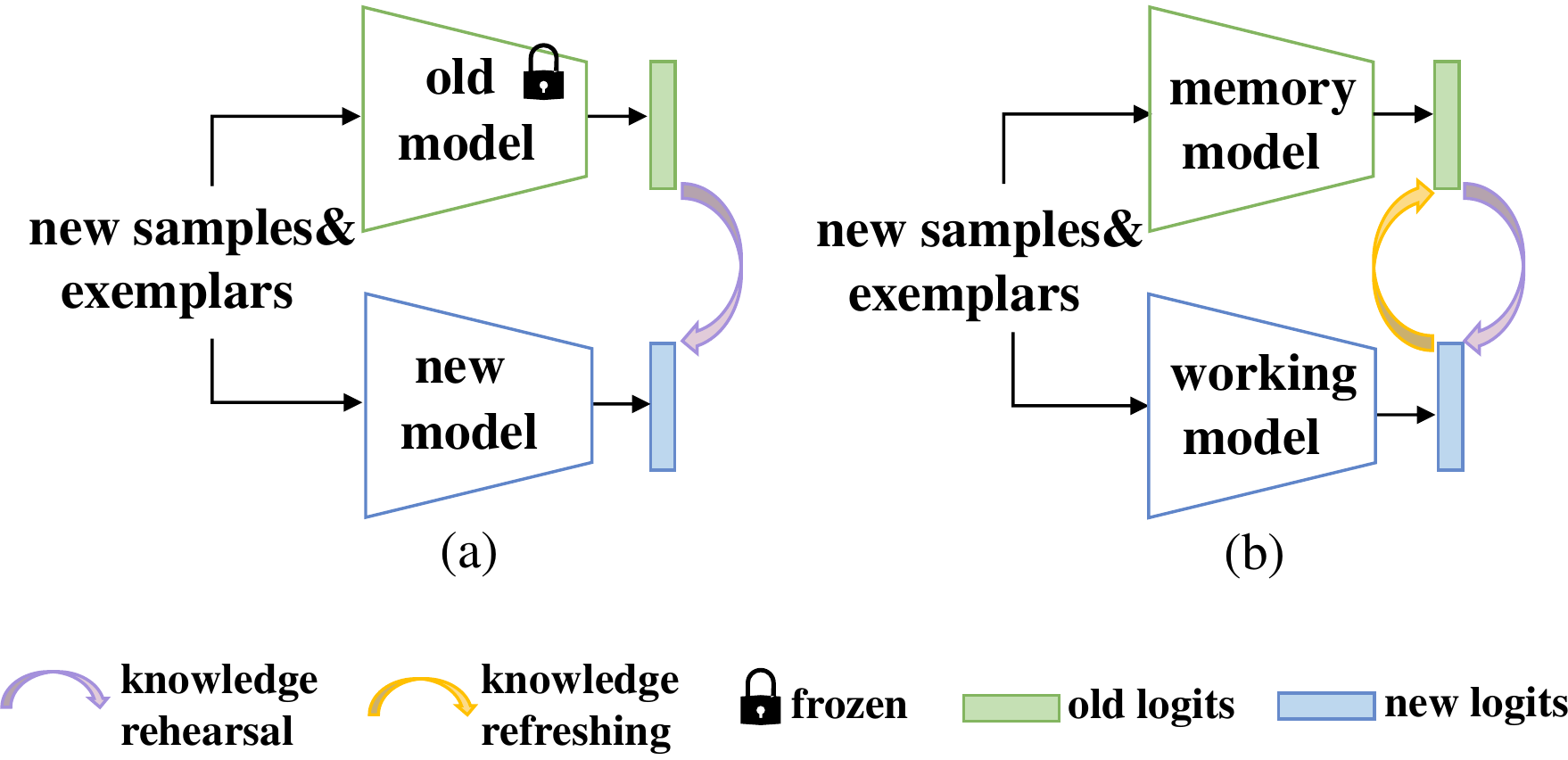}
\caption{Illustration of our method compared with other rehearsal-based methods for LReID. (a) represents a common paradigm of other rehearsal-based approaches that utilize knowledge distillation and rehearsal. (b) represents the centerpiece of our method, which comprises knowledge refreshing with rehearsal and knowledge consolidation.}
\label{fig2}
\vspace{-0.4cm}
\end{figure}
However, pursuing a lifelong model with both positive forward and backward transfer is very challenging. The first challenge is \textit{distribution mismatch}. The natural distribution gaps are prevalent between incrementally added task data. Such sharp distribution gaps will inevitably hinder the efficacy of knowledge transfer between sequential tasks. Unfortunately, previous methods ignore the importance of a smooth transfer, leading to inferior forward transfer on the current task. The second challenge is \textit{capacity mismatch}. Apart from distribution shifts, the incoming datasets may also have non-stationary representation capacities. If so, distillation from a past model will give biased supervision with a weaker generalization ability. This is a hindrance towards both forward and backward transfer.

In light of these challenges, we propose a novel knowledge refreshing and consolidation framework built on top of a knowledge rehearsal baseline. Our design draws inspiration from the human cognitive process, where the somatosensory neocortex and hippocampus coordinate with each other for various functions \cite{sirota2003communication, mizuseki2017hippocampal}. Specifically, our model is equipped with a dynamic memory model (i.e., a hippocampus) and an adaptive working model (i.e., somatosensory neocortex). Then a knowledge refreshing scheme based on knowledge rehearsal operates in an iterative manner to alleviate the negative effects of the distribution shift and capacity mismatch. The interplay between the knowledge rehearsal and the refreshment is based on two considerations. First, the knowledge rehearsal flexibly explores new knowledge without severe forgetting; meanwhile, the knowledge refreshing dynamically calibrates the memory model to avoid harming the performance of the current task, achieving \textit{positive forward transfer}. Second, the knowledge refreshment gradually transfers new insights to the memory model to alleviate the response bias of the memory model, yielding \textit{positive backward transfer}. Further, as the task sequences become longer, the memory model gradually becomes forgetful of past experiences. To this end, we further pave the way for knowledge consolidation from model and feature level to mutually enhance stability. 

To further shed light on our main idea, Fig. \ref{fig2} illustrates the differences between our method and other rehearsal-based methods. Standard rehearsal-based methods, coupled with distillation approaches, are highlighted as an effective solution to lifelong problems, interleaving training samples from the current task with previous exemplars (Fig. \ref{fig2}(a)). Unfortunately, there are several drawbacks in dealing with LReID. As training progresses, the old model is kept static with frozen weights. However, we argue that this strategy is deficient in dynamic retrospection, and therefore prevents the old model from becoming a better teacher. By contrast, our method frees the old model from being unchanged and leaves more space for mutual transfer (Fig. \ref{fig2}(b)). 

To summarize, our contributions are as follows:
\begin{itemize}
\item We propose a novel algorithm named Knowledge Refreshing and Consolidation (KRC) for lifelong person ReID. In addition to dealing with catastrophic forgetting, the proposed KRC also improves model performance on both old and new tasks during the lifelong learning process.

\item Beyond rehearsal, we introduce a knowledge refreshing scheme where the dynamic memory model and the adaptive working model iteratively interact with bi-directional knowledge transfers. Moreover, a knowledge consolidation scheme operating on the dual space is introduced to further improve model stability over the long term.  

\item Extensive experiments demonstrate that KRC outperforms the current state-of-the-art LReID methods by a large margin. Plus, it offers both positive forward transfer and positive backward transfer abilities.

\end{itemize}

\section{Related Work}
\textbf{Lifelong Learning. }
Lifelong learning, also known as continual learning, 
strives to exploit past and current knowledge to adapt a model to new tasks, while maintaining stable performance with old tasks. In general, the research lines of lifelong learning can be grouped into four categories:
regularization-based methods \cite{li2017learning, kirkpatrick2017overcoming, zhao2020maintaining}, 
structure-based methods \cite{serra2018overcoming, yan2021dynamically}, 
algorithm-based methods \cite{lopez2017gradient, guo2022adaptive}.
and rehearsal-based methods \cite{rebuffi2017icarl, hou2019learning, wu2019large}. As our work falls into the rehearsal-based methods, we have concentrated on this stream in this literature review.
Generally, rehearsal-based methods allow one to access a finite number of previous samples, which greatly alleviates the catastrophic forgetting problem. For example, 
UCIR \cite{hou2019learning} exploits cosine normalization and a less-forget constraint to learn a unified feature representation, while BiC \cite{wu2019large} involves a bias correction module to estimate feature drifts in a fully connected layer. WA \cite{zhao2020maintaining} proposes a weight-aligning method, and DER \cite{yan2021dynamically} dynamically adds a new feature extractor per task to expand the representation capability. 
Interestingly, another method GEM reveals that \textit{forward transfer} and \textit{backward transfer} both matter for continual learning. To this end, only a few methods \cite{lopez2017gradient, yan2021dynamically} make some progress in their tasks but are not applicable in LReID.
With this as motivation, the proposed method attempts to pursue a strategy that simultaneously encourages positive forward and backward transfers, all within a limited memory budget.

\textbf{Person Re-Identification.} 
Person ReID has been explored from different perspectives in different settings. 
Fully supervised methods (FS) exploit different architectures and fine-grained features to learn a robust feature representation \cite{li2018harmonious,sun2018beyond,wang2018learning,luo2019strong, he2021transreid, qian2020stripe, Gao_2020_CVPR, chen2021oh}. 
These methods learn a stationary distribution at their best, but often require repeatedly observing the entire prepared dataset. Later approaches investigate unsupervised domain adaptive transfer (UDA) \cite{Wang_2018_CVPR, fu2019self, zheng2021exploiting, zhang2021unsupervised, han2022delving, hu2022divide} or purely unsupervised setups (PU) \cite{yu2019unsupervised, lin2019bottom, wang2020unsupervised, xuan2021intra}. These both seek to use a large corpus of unlabeled images to guide the learning process, either for adaptation purposes or from an unsupervised perspective. While these methods are more readily deployed or flexibly adapted, they usually deliver inferior performance compared with supervised methods. Apart from unsupervised learning, active learning \cite{wang2016human,liu2019deep} provides a human-in-the-loop manner to alleviate the cost of annotation.

Recently, Lifelong ReID has been proposed by AKA \cite{pu2021lifelong} to meet the demand of long-term visual search. This paradigm is imperative for real-world applications, as it requires the ReID system to consolidate knowledge from ever-changing distributions (i.e., newly appearing classes or domains) in a task-incremental fashion. AKA \cite{pu2021lifelong} constructs learnable knowledge graphs which adaptively accumulate knowledge and preserve topologies. GwReID \cite{wu2021generalising}, for example, includes comprehensive learning objectives so as to maintain several coherences during training. PTKP \cite{ge2022lifelong} casts lifelong ReID as a source-free domain adaptation problem, where old tasks are treated as a source domain. This way, the feature space for a new task can be mapped to that of an old task for domain consistency learning. However, when training on a new task, these methods typically focus on retaining old knowledge from a static teacher which may limit the upper bound of performance with the old task \cite{lopez2017gradient}. By contrast, our work not only asks the model to forget less of previous tasks, but it also leaves space for positive knowledge transfer on previous tasks. 

\section{Method}
\subsection{Preliminary}
\subsubsection{Problem Formulation}This section begins by introducing the problem formulation of Lifelong person ReID. Formally, given $N$ sequentially incoming ReID datasets $\mathcal{D} = \{\mathcal{D}^{i}\}_{i=1}^{N}$, we aim to give our ReID model the ability to consolidate sequentially-gained multi-domain knowledge and further encourage positive backward transfer. Specifically, at each step $t$, $\mathcal{D}^{t}=\{\mathcal{D}_{train}^{t},\mathcal{D}_{test}^{t}\}$ contains the training set and testing set. As in previous works, our method builds up an exemplar memory $\mathcal{M}^t$, where a limited number of samples from each previous task are saved.

At training step $t$, we first sample a batch of images $\{(\bm{x}_i^t, y_i^t)\}_{i=1}^{N_b}$ from $\mathcal{D}_{train}^{t}$ and a batch of exemplars $\{(\bm{x}_i^r, y_i^r)\}_{i=1}^{N_b}$ from $\mathcal{M}^{t}$ for replaying, where $\bm{x}_i^t, \bm{x}_i^r$ represent the images, and $y_i^t, y_i^r$ represent the corresponding labels. Plus, a working model $\Theta_t^w$ and a memory model $\Theta_t^m$ are both leveraged for representation learning. In the testing stage of step $t$, the classifier is discarded and the trained feature extractor is used to evaluate on the currently seen domains, which are $\{\mathcal{D}_{test}^{i}\}_{i=1}^{t}$, since the training classes and testing classes are non-overlapping. 
\subsubsection{Method Overview}
 Our method decomposes our learning process into three stages: knowledge rehearsal, knowledge refreshing, and knowledge consolidation. Initially, the first two stages iteratively work together to enable bi-directional knowledge transfer. The knowledge rehearsal allows adaptation of the new tasks and prevents catastrophic forgetting, while the knowledge refreshing scheme mimics the behavior of the knowledge rehearsal mechanism, enabling calibration and memorization of the memory model. When the task is complete, we blend the knowledge of the working model and the memory model in the stage of knowledge consolidation to further enhance stability. In this stage, we exploit the model space consolidation for more stable training, and we explore the feature space consolidation for the generalizable ability.

\begin{figure*}[t]
\centering
\includegraphics[width=\textwidth]{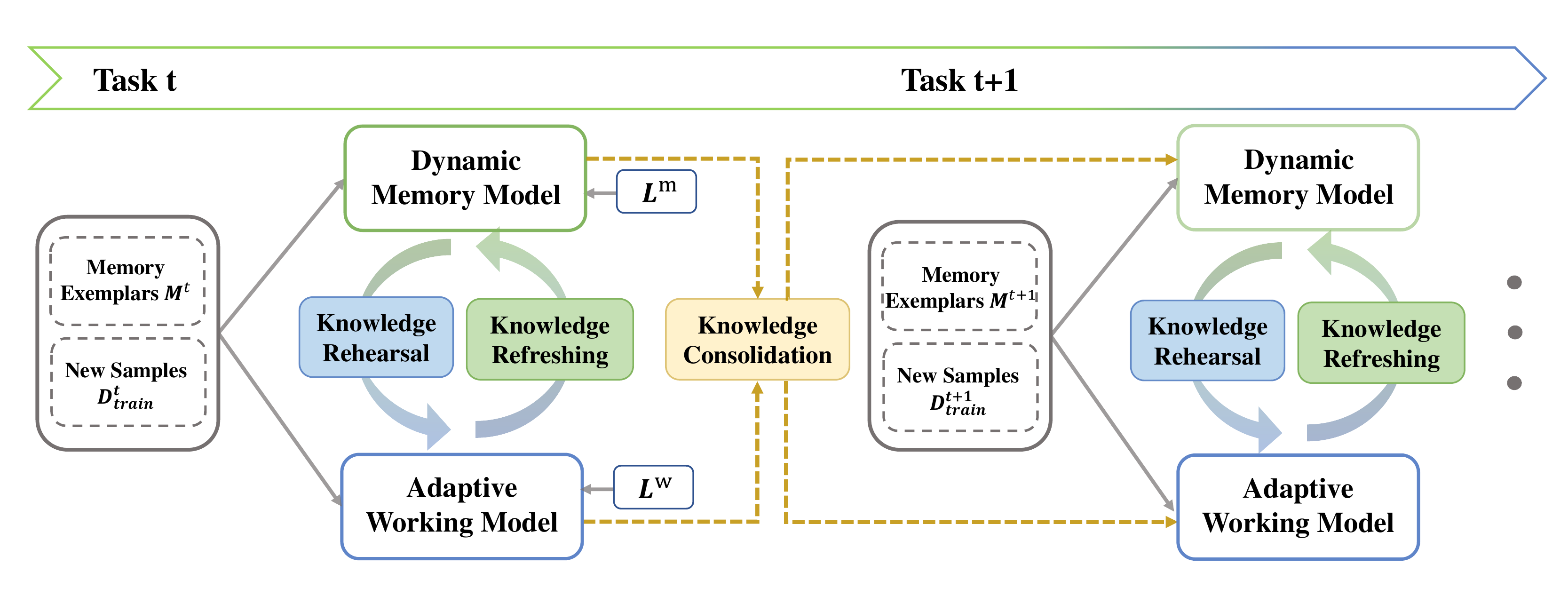}
\caption{An overview of the proposed KRC method for Lifelong person ReID. We first equip our framework with a dynamic memory model and an adaptive working model. Then the memory exemplars and new samples are fed into these models iteratively for knowledge rehearsal and refreshing in bi-directional transfer. After training, the acquired knowledge of these two models is consolidated in preparation for future tasks. The solid lines represent the data flow of both exemplars and new samples, and the dashed lines represent the consolidation process. $\mathcal{L}^{w}$ and $\mathcal{L}^{m}$ denote the overall losses for knowledge rehearsal and knowledge refreshing, respectively.} 
\label{fig3}
\end{figure*}
\subsection{Knowledge Rehearsal}
\;\; In this section, we exploit knowledge rehearsal to allow the model to adapt flexibly to new tasks and to prevent old tasks from being forgotten. 

 To enable knowledge transfer from the memory model to the working model, we use knowledge distillation to minimize the discrepancy between the predictions. Specifically, $\{\bm{p}_i(\cdot)\}_{i=1}^{N_b}$, $\{\bm{q}_i (\cdot)\}_{i=1}^{N_b}$ denote the softmax predictions from the working model and the memory model, taking ($\cdot$) as the input. Here, new samples are exploited for anti-forgetting, which is formulated as:
\begin{small}
\begin{equation}
\mathcal{L}^{w}_{anti}=\frac{1}{N_b}{T^2} \sum_{i=1}^{N_b}\text{JS}(\bm{p}_{i}(\bm{x}_i^t)||\text{SD}(\bm{q}_i(\bm{x}_i^t))),
\label{distw}
\end{equation}
\end{small}where $T$ represents the temperature, $\text{SD}(\cdot)$ represents the stop gradient operation, $\text{JS}(\cdot||\cdot)$ refers to a Jensen-Shannon divergence \cite{fuglede2004jensen}.

Another goal of the knowledge rehearsal is to make the adaptation flexible. Therefore, a loss for adaptation is needed, which contains a cross-entropy loss and a triplet loss. Here, the cross-entropy loss is used on new samples, so we have:
\begin{small}
\begin{equation}
\mathcal{L}_{ce}^{w}=-\frac{1}{N_b}\sum_{i=1}^{N_b}\log\bm{p}_i(\bm{x}_i^t)^{(y_i^t)},
\label{cew}
\end{equation}
\end{small}where $\bm{p}_i(\bm{x}_i^t)^{(y_i^t)}$ is the $y_i^t$-th digit of the softmax prediction $\bm{p}_i(\bm{x}_i^t)$. Meanwhile, we optimize the working model on both current samples and exemplars for metric learning, where the triplet loss is defined as:
\begin{small}
\begin{equation}
\begin{aligned}
\mathcal{L}_{trip}^{w} = &\frac{1}{N_b}\sum_{i=1}^{N_b} \max(d(\bm{a}_i^{w,t}, \bm{p}_i^{w,t}) - d(\bm{a}_i^{w,t}, \bm{n}_i^{w,t}) + m, 0) \\&+\frac{1}{N_b}\sum_{i=1}^{N_b} \max(d(\bm{a}_i^{w,r}, \bm{p}_i^{w,r}) - d(\bm{a}_i^{w,r}, \bm{n}_i^{w,r}) + m, 0),
\label{tripw}
\end{aligned}
\end{equation}
\end{small}where  $d(\cdot, \cdot)$ represents the euclidean distance and $m$ is the margin; ($\bm{a}_i^{w,t}, \bm{p}_i^{w,t}, \bm{n}_i^{w,t}$) and ($\bm{a}_i^{w,r}, \bm{p}_i^{w,r}, \bm{n}_i^{w,r}$) are two triplets of anchor sample, positive sample and negative sample of $\bm{x}_i^{t}$ and $\bm{x}_i^{r}$, respectively. Here, the triplets are extracted from the working model. 

Now the working model is equipped with two learning objectives and we get the overall knowledge rehearsal loss $\mathcal{L}^w$ to update the working model:
\begin{equation}
\label{krh}
\mathcal{L}^w = \mathcal{L}_{anti}^{w} +\mathcal{L}_{ce}^{w} +\mathcal{L}_{trip}^{w}
\end{equation}

It is worth noting that, to enable positive forward transfer, exemplars for the cross-entropy loss and the distillation loss are excluded, and only exploited for the triplet loss. This works to avoid over-confident predictions and supervision of exemplars given by the softmax functions. As such, the model can more flexibly adapt to new domains while preserving old knowledge at the same time.

\subsection{Knowledge Refreshing}
Whereas distillation-based knowledge rehearsal methods explore learned experiences from the memory model, such experiences could be biased. For this reason, distilling two sharply different distributions may not be very effective if the working model veers towards the current task. By contrast, our approach leverages the new identities in the current task as negative classes to update the memory model in a slow fashion, inspiring novel insights and refreshing the old knowledge. More specifically, we simply propose a knowledge refreshing scheme that mirrors the knowledge rehearsal on a similar note, however, with a smaller learning rate, which turns the memory model into a dynamic teacher.

Our first goal is to calibrate the responses of the memory model. Specifically, we use a distillation loss that is similar to $\displaystyle \mathcal{L}_{anti}^{w}$ but we update the memory model in a smooth manner, which is different from previous practice. This way, the responses of the memory model can be better calibrated. Thus, $\mathcal{L}_{cali}^{m}$ for memory calibration is formulated as:
\begin{small}
\begin{equation}
\mathcal{L}^{m}_{cali}=\frac{1}{N_b}{T^2} \sum_{i=1}^{N_b}\text{JS}(\bm{q}_{i}(\bm{x}_i^t)||\text{SD}(\bm{p}_i(\bm{x}_i^t))),
\label{distm}
\end{equation}
\end{small}

In addition, we also enhance the ability of the memory model to memorize various patterns of person identities. Specifically, we include a cross-entropy loss on new samples to update the memory model:
\begin{small}
\begin{equation}
\mathcal{L}_{ce}^{m}=-\frac{1}{N_b}\sum_{i=1}^{N_b}\log\bm{q}_i(\bm{x}_i^t)^{(y_i^t)},
\label{cem}
\end{equation}
\end{small}Then we add a triplet loss on the memory model. We define ($\bm{a}_i^{m,t}, \bm{p}_i^{m,t}, \bm{n}_i^{m,t}$) and ($\bm{a}_i^{m,r}, \bm{p}_i^{m,r}, \bm{n}_i^{m,r}$) as the triplets of anchor sample, positive sample and negative sample of $\bm{x}_i^t$ and $\bm{x}_i^r$. Both of these triplets are extracted from the memory model. Thus, $\mathcal{L}_{trip}^m$ is formulated as:
\begin{small}
\begin{equation}
\begin{aligned}
\mathcal{L}_{trip}^{m} = &\frac{1}{N_b}\sum_{i=1}^{N_b} \max(d(\bm{a}_i^{m,t}, \bm{p}_i^{m,t}) - d(\bm{a}_i^{m,t}, \bm{n}_i^{m,t}) + m, 0) \\& + \frac{1}{N_b}\sum_{i=1}^{N_b} \max(d(\bm{a}_i^{m,r}, \bm{p}_i^{m,r}) - d(\bm{a}_i^{m,r}, \bm{n}_i^{m,r}) + m, 0),
\end{aligned}
\label{tripm}
\end{equation}
\end{small}

Through calibration and memorization, we cast the memory model as a dynamic teacher to offer more meaningful guidance. Therefore, the memory model is optimized through the overall knowledge refreshing loss $\mathcal{L}^m$, which is formulated as:
\begin{equation}
\label{krf}
\mathcal{L}^{m} = \mathcal{L}_{cali}^{m} +\mathcal{L}_{ce}^{m} + \mathcal{L}_{trip}^{m},
\end{equation}

We remark that: (i) Similar to the knowledge rehearsal scheme, the exemplar memory only participates in metric learning, since we empirically find the cross-entropy loss or logit-level distillation on $\mathcal{M}^{t}$ may possibly put the ReID model at risk of overfitting on the exemplars. This would, of course, harm the performance to generalize on previous ReID tasks. (ii) While the learning objectives are similar to those of the knowledge rehearsal scheme, there is a different motivation for this stage.  With these learning objectives, the memory model was better able to communicate with the working model and could then refresh itself through gradual adaptation. (iii) The knowledge rehearsal stage and the refreshing stage iteratively coordinate with each other, mimicking the information flow between the hippocampal–entorhinal loop during the sleep/wake cycle \cite{mizuseki2017hippocampal}. The structure of our algorithm is given in Alg. \ref{alg:algorithm}.

\begin{algorithm}[t]
\caption{KRKC}
\label{alg:algorithm}
\textbf{Input}: Incoming datasets $\mathcal{D}_{train}^{t}$, Buffer $\mathcal{M}^{t}$, Learning rate $\gamma$, $\eta$.\\
\textbf{Parameter}: Parameters of working model: $\Theta^{w}_t$; Parameters of memory model: $\Theta^{m}_t$.

\begin{algorithmic}[1] 
\FOR{epoch in $1\rightarrow e_{max} $}
    \FOR{\{$\bm{x}_i^t, y_i^t\}_{i=1}^{N_b}$ \textbf{in} $\mathcal{D}_{train}^{t}$}
    \STATE Sample a mini-batch $\{\bm{x}_i^r, y_i^r\}_{i=1}^{N_b}$ from $\mathcal{M}^{t}$.
    \STATE Knowledge rehearsal for anti-forgetting by Eq. (\ref{distw}).
    \STATE Knowledge rehearsal for adaptation by Eq. (\ref{cew}), Eq. (\ref{tripw}).
    \STATE Calculate overall loss $\mathcal{L}^{w}$ by Eq. (\ref{krh}).
    \STATE Update $\Theta^{w}_t$ by gradient descent:
    \STATE $\vartheta\leftarrow \vartheta -\gamma\nabla_{\vartheta}\mathcal{L}^w$ \textbf{for} $\vartheta$ \textbf{in} $\Theta_{t}^{w}$.
    \STATE Knowledge refreshing for calibration by Eq.(\ref{distm}).
    \STATE Knowledge refreshing for memorization by Eq.(\ref{cem}), Eq. (\ref{tripm}).
    \STATE Calculate overall loss $\mathcal{L}^m$ by Eq. (\ref{krf}).
    \STATE Update $\Theta^{m}_t$ by gradient descent:
    \STATE $\vartheta\leftarrow \vartheta -\eta\nabla_{\vartheta}\mathcal{L}^m$ \textbf{for} $\vartheta$ \textbf{in} $\Theta^{m}_t$.
    \ENDFOR
\ENDFOR
\STATE Model space consolidation by Eq. (\ref{msc}).
\STATE Feature space consolidation by Eq. (\ref{fsc}).
\STATE Update exemplar memory: $ \mathcal{M}^{t}\rightarrow\mathcal{M}^{t+1}$.
\STATE \textbf{return} updated model parameters $\Theta^{w}_{t+1}$, $\Theta^{m}_{t+1}$.
\end{algorithmic}
\end{algorithm}

\subsection{Knowledge Consolidation}
\;\; To further combat catastrophic forgetting and improve the model's generalization ability for long task sequences, we carefully consolidated the old and the new knowledge in the model space and the feature space, with the purpose of retaining a minimum for all tasks.
\subsubsection{Model Space Consolidation}
Normally, memory models play a key role in distilling old knowledge into working models. However, the memory model is discarded immediately after training the current task, and it is not involved in subsequent tasks. To account for this disparity, at the end of each step, our working model is reorganized with the parameters of the memory model to linearly combine their merits. Intuitively, greater importance is placed on the memory model as the task number increases since the memory model comprises richer knowledge of previous tasks in the long run. Formally, the update rule is defined as:
\begin{equation}
\label{msc}
\begin{aligned}
\Theta^{w}_{t+1} &= \frac{1}{t+1} \Theta^{w}_{t}+ \frac{t}{t+1} \Theta^{m}_{t},
\end{aligned}
\end{equation}
Next, a copy of $\Theta^{w}_{t+1}$ is retained to initialize $\Theta^{m}_{t+1}$ for the next task.
\subsubsection{Feature Space Consolidation} Thanks to the knowledge rehearsal and refreshment mechanism, we now have two models that perform comparably well on the current task and previous tasks, but with different focuses. To take full advantage of the working model and the memory model, we devise a simple but effective way to consolidate the feature space, which is also important for generalization ability. Specifically, once the task is complete, we have two features $\bm{f}^{w}_{t}$,$\bm{f}^{m}_{t}$ from the trained feature extractors $\Theta^{w}_t$ and $\Theta^{m}_t$. These two features are concatenated and the fused feature is used for person retrieval, which is tantamount to:
\begin{equation}
\label{fsc}
\bm{f}^{fuse}_{t} = [\bm{f}^{w}_{t};\bm{f}^{m}_{t}].
\end{equation}
\begin{table*}
\centering
	\caption{Performance comparison of the state-of-the-art methods on seen domains. All the reported results are implemented with the released code on our baseline. The training order is VIPeR $\rightarrow$ Market $\rightarrow$ CUHK-SYSU $\rightarrow$ MSMT17.}
	\begin{tabular}{c|cc|cc|cc|cc|cccccccccc}
        \toprule[1.1pt]
        \multirow{2}{*}{\textbf{Method}} 
		&\multicolumn{2}{c}{\textbf{VIPeR}}
		&\multicolumn{2}{c}{\textbf{Market}}
		&\multicolumn{2}{c}{\textbf{CUHK-SYSU}}
        &\multicolumn{2}{c}{\textbf{MSMT17}}
        &\multicolumn{2}{c}{\textbf{Average}}\\
        \cmidrule{2-11}
		 &\textbf{R-1} &\textbf{mAP} &\textbf{R-1}	&\textbf{mAP} &\textbf{R-1}	&\textbf{mAP} &\textbf{R-1} &\textbf{mAP} &\textbf{$\overline{s}_{\text{R-1}}$} &\textbf{$\overline{s}_{\text{mAP}}$} \\
		\midrule\midrule
		Joint  &78.5 &85.7 &93.7 &85.4 &89.8&88.1 &74.7 &49.4 &84.2 &77.2\\ 
        \midrule
		LwF \cite{li2017learning} &45.3 &54.6 &74.6 &53.9 &83.7 &81.1 &32.3 &14.6 &59.0 &51.1\\ 
        iCarL \cite{rebuffi2017icarl} &56.6 &67.0 &78.7 &56.9 &82.9&80.0 &24.1 &10.2&60.6&53.6\\
        UCIR \cite{hou2019learning}  &56.3 &66.8 &65.9 &45.5 &70.5 &68.8 &29.3 &12.8&55.5&48.5\\
		BiC \cite{wu2019large} &50.4 &61.2 &68.9 &47.4 &72.5 &71.6 &42.1 &22.4&58.5&50.7\\
		WA \cite{zhao2020maintaining} &48.1 &58.0 &70.3 &50.9 &71.9 &70.4 &38.8 &18.6&57.2&49.5\\
		\midrule
	   AKA \cite{pu2021lifelong} &50.6&61.7 &50.7 &28.3 &79.6 &76.9 &28.0 &13.4&52.2&45.1\\
		PTKP \cite{ge2022lifelong} &56.0 &66.3 &77.4 &58.3 &78.8 &77.1 &48.0&25.2&65.1&56.7\\
		\midrule
	 Ours  &\textbf{67.7} &\textbf{76.3} &\textbf{82.5} &\textbf{64.4} &\textbf{90.7} &\textbf{88.9} &\textbf{67.1} &\textbf{43.3} &\textbf{77.0}&\textbf{68.2}\\
    	\bottomrule[1.1pt]
	\end{tabular}
	\label{maintab}
\end{table*}
\begin{table}
        \centering
	\caption{Performance comparison of the state-of-the-art methods over unseen domains. The results (mAP) are given by trained models at the last step.}
	\begin{tabular}{c|c|c|c|c}
        \toprule[1.1pt]
        
       \textbf{Method} &\textbf{CUHK01} &\textbf{CUHK02} &\textbf{CUHK03} &\textbf{PRID}\\
       \midrule\midrule
       Joint &76.6 &68.8 &31.9 &36.6\\ \midrule
       LwF&60.1&52.0&25.0&27.0\\ 
       iCaRL&60.4 &55.1 &20.8&34.9\\ 
       UCIR&40.7&37.6&21.7&22.4\\ 
       BiC&50.4&44.6&31.2&24.1\\ 
       WA&50.2&48.5&29.6&21.9\\ 
       \midrule
       AKA&44.3&42.1&21.2&22.7\\
       PTKP&57.4&52.3&30.6&28.4\\ \midrule
       Ours &\textbf{79.2}&\textbf{69.8}&\textbf{39.5}&\textbf{52.7}\\
       \bottomrule[1.1pt] 
	\end{tabular}
    \vspace{-15pt}
\label{unseen}
\end{table}

\section{Experiment}
\subsection{Datasets and Evaluation Protocal}
\;\; To evaluate our strategy's performance with LReID tasks, we assess our model on a challenging benchmark with four sequential datasets: VIPeR \cite{gray2008viewpoint}, Market \cite{zheng2015scalable}, CUHK-SYSU \cite{xiao2016end}, and MSMT17 \cite{wei2018person}. These datasets mimic real-world scenarios. With CUHK-SYSU, we follow the same data pre-processing for training and testing as in PTKP \cite{ge2022lifelong}. Then, to further evaluate our method's generalization ability on unseen datasets, we compare a range of methods on the CUHK01 \cite{li2012human}, CUHK02 \cite{li2013locally}, CUHK03 \cite{li2014deepreid}, and PRID \cite{hirzer2011person} benchmarks. 

The evaluation metrics for each task include standard mean Average Precision (mAP) and Rank-1 accuracy (R-1). For the lifelong evaluations on the seen tasks, we report the ``Average" accuracy which is the average incremental accuracy of mAP/Rank-1 after the last step \cite{ge2022lifelong}.

\subsection{Implementation Details}
\;\; We use a ResNet50 model pre-trained on ImageNet as our backbone. Global average pooling is replaced with generalized mean pooling. All person images are resized to $\text{256}\times \text{128}$. The batch size for the current task and the exemplar task is set to 128. For the memory update, we follow PTKP \cite{ge2022lifelong}, which saves the farthest two samples for each identity, with 250 identities per task. We use Adam for optimization and train each task for 60 epochs. The learning rate is initialized at $\text{3.5}\times{\text{10}^{-4}}$, which is then decreased by $\text{0.1}$ after the $\text{40}^{th}$ epoch for the first task. The learning rate in the knowledge refreshing scheme is initialized as $\text{3.5}\times{\text{10}^{-5}}$ for subsequent tasks and $\text{3.5}\times{\text{10}^{-6}}$ for the last task, all decreased by $\text{0.1}$ after the $\text{30}^{th}$ epoch. The initial learning rate of the knowledge rehearsal mechanism is set to $\text{3.5}\times{\text{10}^{-4}}$, with the other settings kept the same as the knowledge refreshing scheme. The weights for all losses are set to 1 and the temperature $T$ for distillation is set to 2.
\subsection{Comparison with the State-of-the-Arts}
\subsubsection{Results on Seen Domains}
Table \ref{maintab} summarizes the results of LReID benchmark on seen domains. As shown, our method consistently outperforms the other rehearsal-based methods and LReID methods by a clear margin, with an average gain of +11.9\% w.r.t. the average incremental R-1 accuracy ($\overline{s}_{\text{R-1}}$) and +11.5\% w.r.t. the average incremental mAP accuracy ($\overline{s}_{\text{mAP}}$), respectively. Additionally, our method adds more plasticity to new tasks, achieving 67.1\% Rank-1 and 43.3\% mAP on MSMT with at least +19.1\% performance boost. We argue that this owes to our knowledge refreshing scheme which unleashes sufficient flexibility for further exploration of new tasks. 
\begin{table*}[htbp]
	\centering
	\begin{tabular}{cccc||cc|cc|cc|cc|cccc}
		\toprule[1.1pt]
        \multirow{2}{*}{\textbf{KRH}} & \multirow{2}{*}{\textbf{KRF}}  &\multirow{2}{*}{\textbf{MSC}}&\multirow{2}{*}{\textbf{FSC}}&\multicolumn{2}{c|}{\textbf{VIPeR}}
		&\multicolumn{2}{c|}{\textbf{Market}}
		&\multicolumn{2}{c|}{\textbf{CUHK-SYSU}}
        &\multicolumn{2}{c|}{\textbf{MSMT17}}
        &\multicolumn{2}{c}{\textbf{Average}}\\
        \cmidrule{5-14}
        &&&&\textbf{R-1} &\textbf{mAP} &\textbf{R-1} &\textbf{mAP} &\textbf{R-1}	&\textbf{mAP} &\textbf{R-1} &\textbf{mAP}&\textbf{$\overline{s}_{\text{R-1}}$}&\textbf{$\overline{s}_{\text{mAP}}$}\\
		\midrule
		\checkmark &&&&49.4&59.8&62.0&37.2&81.8&78.5&30.1&12.2&55.8&46.9 \\
		\checkmark &\checkmark&	 &&57.0	&67.8 &72.1&47.6&87.3&85.4&66.4&42.9&70.7&60.9\\
        \checkmark & &\checkmark &\checkmark&60.1&70.0&72.7&51.3&84.6&82.5&34.4&14.8&63.0&54.7\\
		\checkmark &\checkmark&\checkmark&&58.5 &69.9&72.3&47.9&86.4&84.4&69.9 &46.1&71.8&62.1\\
  \checkmark &\checkmark &&\checkmark&61.7&71.9&81.7&61.6&91.4&89.6&59.0&36.0&73.5&64.8\\
  \checkmark &\checkmark&	\checkmark &\checkmark &\textbf{67.7} &\textbf{76.3} &\textbf{82.5} &\textbf{64.4} &\textbf{90.7} &\textbf{88.9} &\textbf{67.1} &\textbf{43.3} &\textbf{77.0}&\textbf{68.2}\\
  
		\bottomrule[1.1pt]
	\end{tabular}
    \caption{Ablation study of individual component analysis. We reported the accuracy of four tasks at the last training step.}
	\label{ablation}
\end{table*}
\subsubsection{Results on Unseen Domains}
Table \ref{unseen} summarizes the results of generalization on unseen domains. Our method reaches a more stable result over four unseen datasets. Compared with other works, our method even outperforms joint training on CUHK01, CUHK03, and PRID. Specifically, our method surpasses joint training by +13.5\% on PRID, which demonstrates that our method does enable a smooth transfer and robust generalization ability over various unseen datasets. 

\subsection{Ablation Study} 
\subsubsection{Individual Component Analysis} 
To better understand the contribution of
each module and how these modules corroborate each other, we reported the results w.r.t. the accuracy of four tasks at the last training step. From the results in Table \ref{ablation}, we made the following interpretations. First, on top of knowledge rehearsal, the knowledge refreshing scheme (KRF) consistently and gradually adds new insights to the memory model, which in turn helps to learn the working model. This is an indispensable part of maximally encouraging positive backward transfer for previous tasks. Second, knowledge consolidation (KC) comprehensively equips our model with stability and resilience against ever-increasing task sequences. Specifically, model space consolidation (MSC) and feature space consolidation (FSC) both mitigate catastrophic forgetting. More importantly, our model attains significant performance gains when these two modules work together. Further, knowledge consolidation (KC) and knowledge refreshing (KR) can complement each other as their collaboration considers both short-term plasticity and long-term stability.

\subsection{Stability and Plasticity Analysis}
\subsubsection{Positive Backward Transfer with Stability}
To assess the ability to use new data to improve performance on previous tasks, we introduce a memory refreshing scheme to gradually absorb new knowledge and encourage a smooth transfer, which helps preclude the response bias of the memory model. Formally, we define $R_{i,j}$ as the test accuracy on task $j$ after observing task $i$ and $T$ as the total number of tasks. At each step $i$, we can observe the performance change of task $j$ from $R_{j,j}$ to $R_{i,j}$. By averaging these local accuracy changes we compute the overall backward  transfer value (BWT) of all previous tasks as:
\begin{equation}
\text{BWT} = \frac{1}{T-1}\sum_{i}^{T-1}\frac{1}{i}\sum_{j=1}^{i} (R_{i,j} -R_{j,j})
\end{equation}
As seen in Table \ref{transfer}, our method gives +5.3\% positive backward transfer.
\subsubsection{Positive Forward Transfer with Plasticity}
To measure the influence of previous knowledge on the performance of new tasks, we also evaluate the forward transfer ability (FWT), which is defined by:
\begin{equation}
\text{FWT} = \frac{1}{T-1}\sum_{i=2}^{T} (R_{i,i} - \overline{R}_{i,i})
\end{equation}
where $\overline{R}_{i,i}$ represents the accuracy given by the model trained at random initialization. As shown in Table \ref{transfer}, most methods sacrifice plasticity to maintain stability, achieving negative forward transfer. Nevertheless, our method achieves +3.9\% positive forward transfer in R-1 accuracy, which benefits from our KRC module that forms a closed-loop positive feedback system for subsequent tasks.
\begin{table}
	\begin{tabular}{c|cccccccc}
        \toprule[1.1pt]
       \textbf{Method} &\textbf{LwF} &\textbf{iCaRL} &\textbf{WA} &\textbf{AKA} &\textbf{PTKP} &\textbf{Ours}\\
       \midrule
       BWT &-0.9 &3.7 &-0.5 &-1.4&3.4&\textbf{5.3}\\ \midrule
       FWT&-14.6&-19.1&-17.6&-22.4&-13.2&\textbf{3.9}\\ 
       \bottomrule[1.1pt]
    \end{tabular}
    \caption{Backward transfer (BWT) and forward transfer (FWT) measured in R-1 accuracy.}
\label{transfer}
\end{table}

\begin{figure}[t]
	\centering
		\subfloat[]{
		\includegraphics[width=0.49\linewidth]{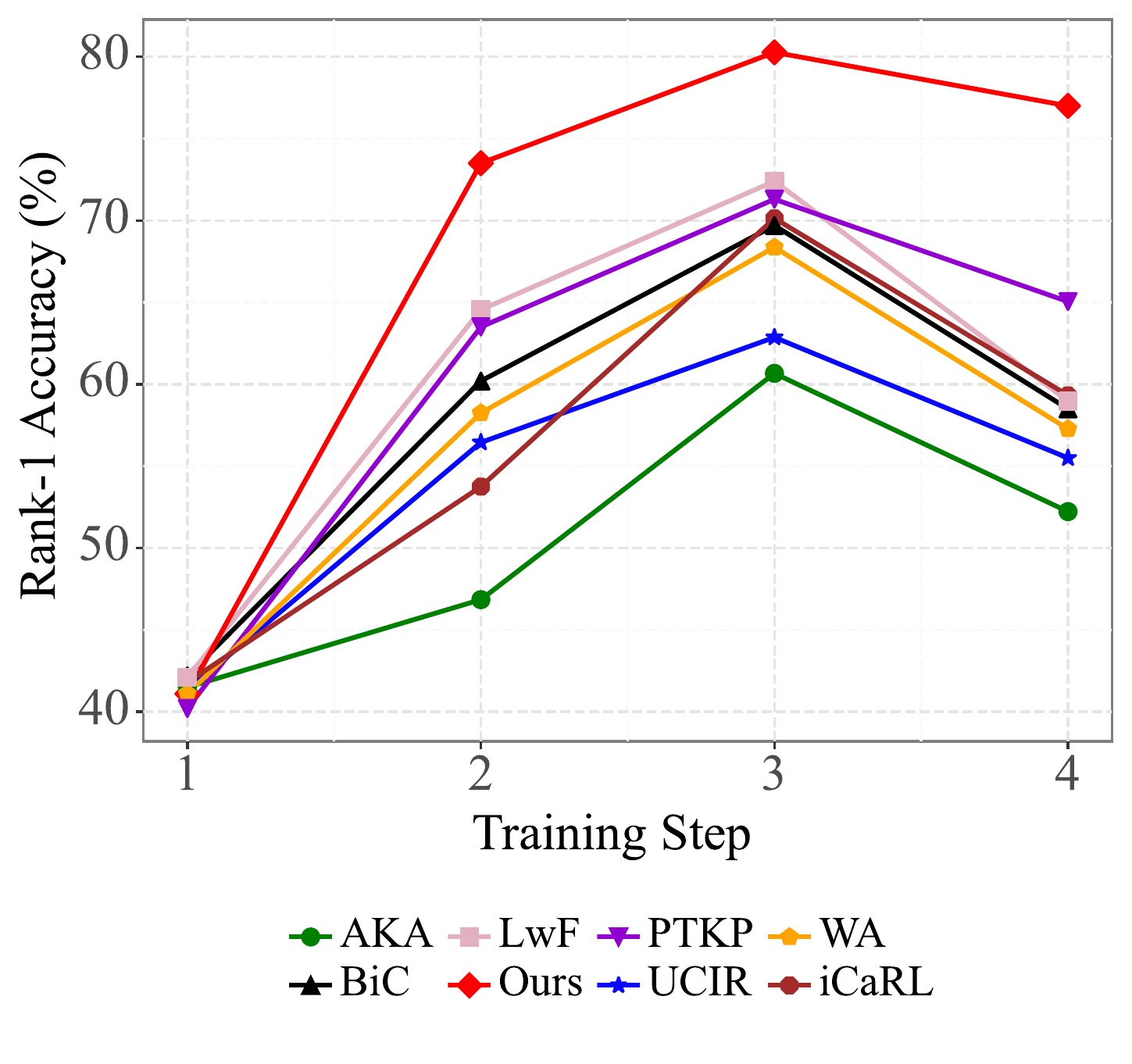}
		\label{1a}
	}
	\subfloat[]{
		\includegraphics[width=0.49\linewidth]{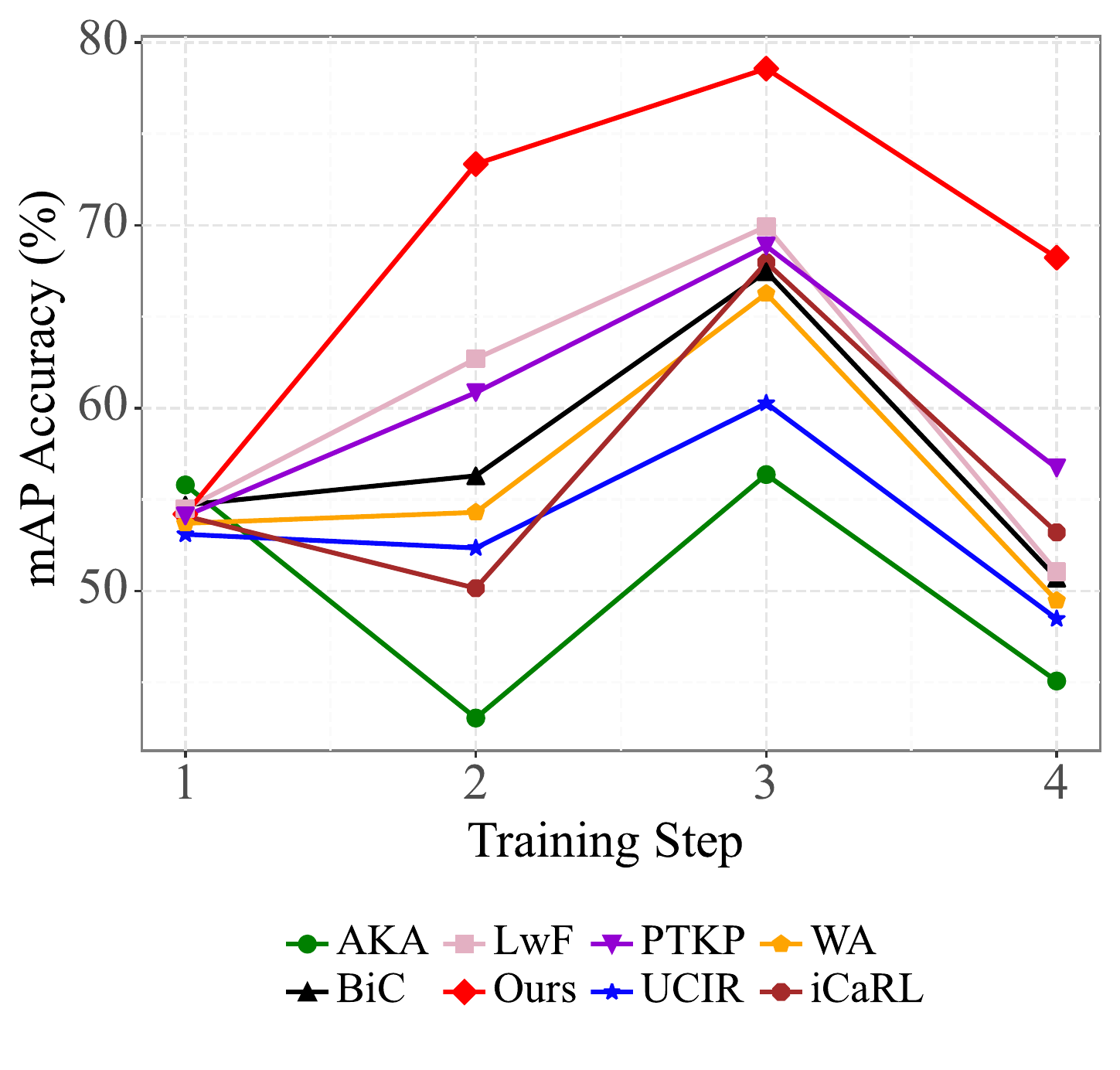}
		\label{1b}
	}

\caption{(a) Average incremental accuracy w.r.t. $\overline{s}_{\text{R-1}}$. (b) Average incremental accuracy w.r.t. $\overline{s}_{\text{mAP}}$.}
\label{vis}
\end{figure}
\section{Conclusion}
In this paper, we propose a novel rehearsal-based framework that not only serves the purpose of anti-forgetting, but also enables both positive forward and backward transfer. To meet this goal, we introduce a knowledge refreshing scheme to iteratively transfer knowledge between the memory and the working modules. Moreover, the knowledge consolidation scheme further improves model stability in the long term. Extensive experiments are conducted on seen and unseen domains. The performance gains over the state-of-the-art methods demonstrate that our method achieves both positive forward and backward transfer for LReID.
\section{Acknowledgement}
This work was supported by the Shanghai Sailing Program (21YF1429400, 22YF1428800), Shanghai Frontiers Science Center of Human-centered Artificial Intelligence (ShangHAI), National Key R\&D Program of China (2018AAA0100704), NSFC (61932020, 62172279), Science and Technology Commission of Shanghai Municipality (20ZR1436000), Program of Shanghai Academic Research Leader, and Shuguang Program supported by Shanghai Education Development Foundation.
\bibliographystyle{aaai23}
\bibliography{aaai23}

\end{document}